\documentclass[letterpaper, 10 pt, conference]{ieeeconf}  
\usepackage{amsmath}
\usepackage{amssymb}

\usepackage{times}
\usepackage{epsfig}
\usepackage{graphicx}
\usepackage{amsmath}
\usepackage{mathabx}
\usepackage{bm}
\usepackage{amssymb}

\usepackage{tabularx}
\usepackage{amsmath}
\usepackage{siunitx}
\usepackage{booktabs}
\usepackage{multirow}
\usepackage[table,xcdraw]{xcolor}
\usepackage{float}
\usepackage[normalem]{ulem}
\useunder{\uline}{\ul}{}
\newcommand{\etal}{et al.}
\usepackage{cite}
\usepackage{booktabs}
\usepackage{caption}
\usepackage{subcaption}
\usepackage{hyperref}

\IEEEoverridecommandlockouts                              

\title{\LARGE \bf
Adversarial Attacks on Monocular Pose Estimation
}

\author{Hemang Chawla, Arnav Varma, Elahe Arani*\thanks{*Equal advising.}, and Bahram Zonooz*
\thanks{All authors are with Advanced Research Lab, NavInfo Europe, The Netherlands.
Contact: \tt\small hemang.chawla@navinfo.eu}%
}

\begin{document}

\maketitle
\thispagestyle{empty}
\pagestyle{empty}

\begin{abstract}
Advances in deep learning have resulted in steady progress in computer vision with improved accuracy on tasks such as object detection and semantic segmentation. Nevertheless, deep neural networks are vulnerable to adversarial attacks, thus presenting a challenge in reliable deployment. Two of the prominent tasks in 3D scene-understanding for robotics and advanced drive assistance systems are monocular depth and pose estimation, often learned together in an unsupervised manner. While studies evaluating the impact of adversarial attacks on monocular depth estimation exist, a systematic demonstration and analysis of adversarial perturbations against pose estimation are lacking. We show how additive imperceptible perturbations can not only change predictions to increase the trajectory drift but also catastrophically alter its geometry. We also study the relation between adversarial perturbations targeting monocular depth and pose estimation networks, as well as the transferability of perturbations to other networks with different architectures and losses. Our experiments show how the generated perturbations lead to notable errors in relative rotation and translation predictions and elucidate vulnerabilities of the networks. 
\footnote{Code can be found at \href{https://github.com/NeurAI-Lab/mono-pose-attack}{https://github.com/NeurAI-Lab/mono-pose-attack}.}
\end{abstract}

\section{Introduction}
Vision systems have benefited immensely from the progress in deep neural networks enabling a wide variety of scene-understanding tasks such as object detection and semantic segmentation. Recently, inspired by the concepts in structure-from-motion, methods that simultaneously train neural networks to predict scene-depth and camera pose in an unsupervised manner have also been proposed~\cite{packnet, bian2021unsupervised, varma2022transformers}. Such 3D scene-geometry tasks have a significant role in planning and navigation of robots, self-driving cars, and robot-assisted surgeries~\cite{pillai2017towards, huang2021self}. Nonetheless, while deep neural networks have achieved state-of-the-art performance in several perception tasks, they remain vulnerable to adversarial attacks that fool the networks with often imperceptible changes to the input. This limits their applicability in safety-critical tasks where the margin for error is low. 



\begin{figure}[t]
\centering
\begin{subfigure}{\linewidth}
  \centering
  \includegraphics[width=0.9\linewidth]{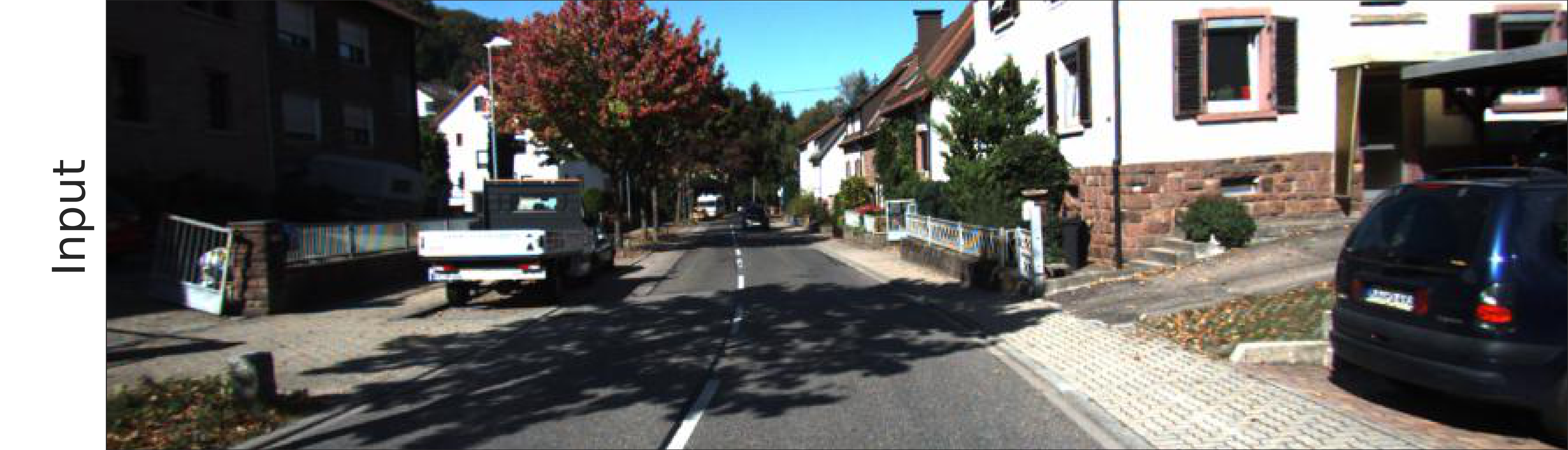} 
\end{subfigure}

\medskip

\begin{subfigure}{\linewidth}
  \centering
  \includegraphics[width=0.9\linewidth]{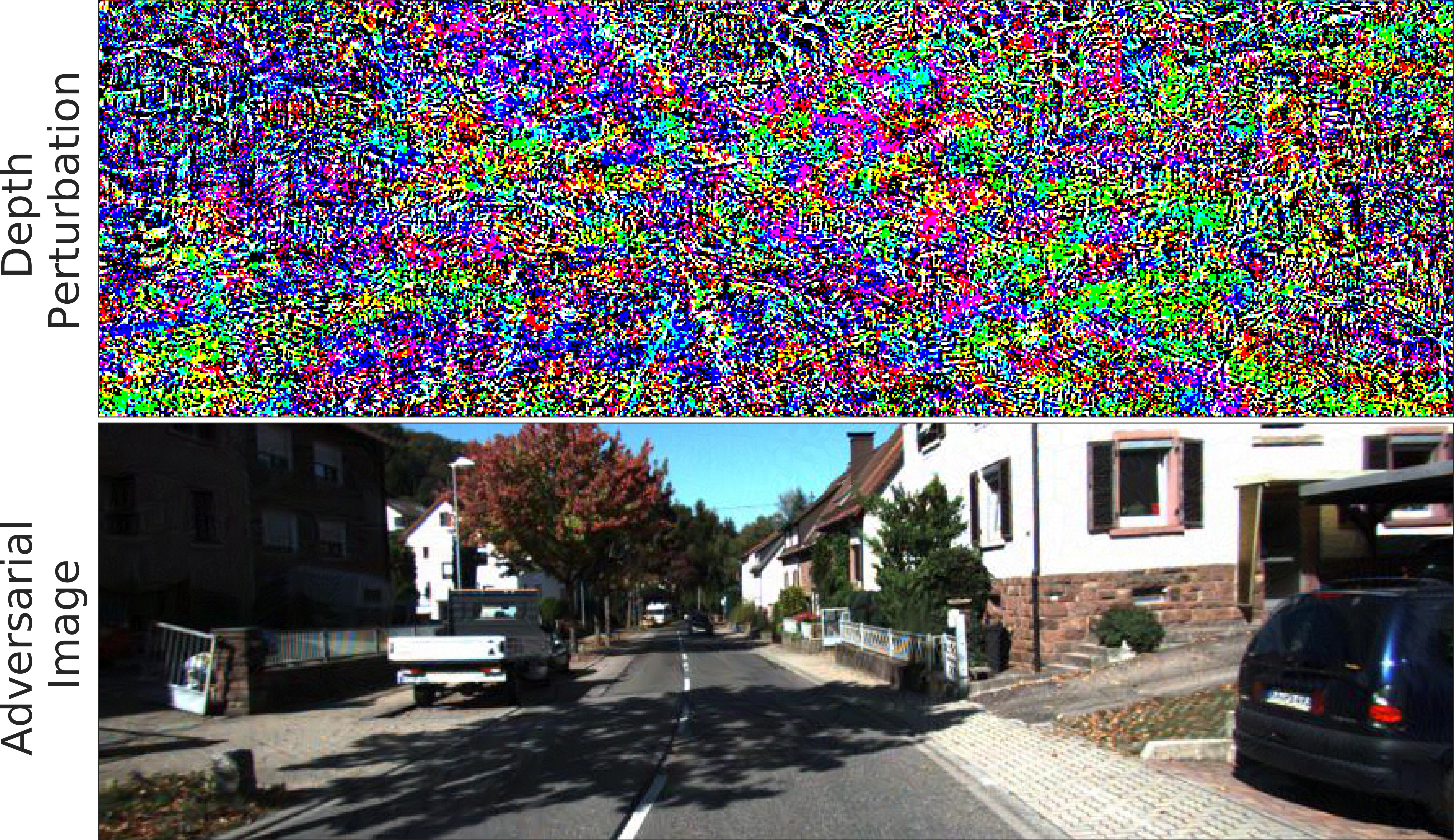} 
\end{subfigure}

\medskip

\begin{subfigure}{\linewidth}
  \centering
  \includegraphics[width=0.9\linewidth]{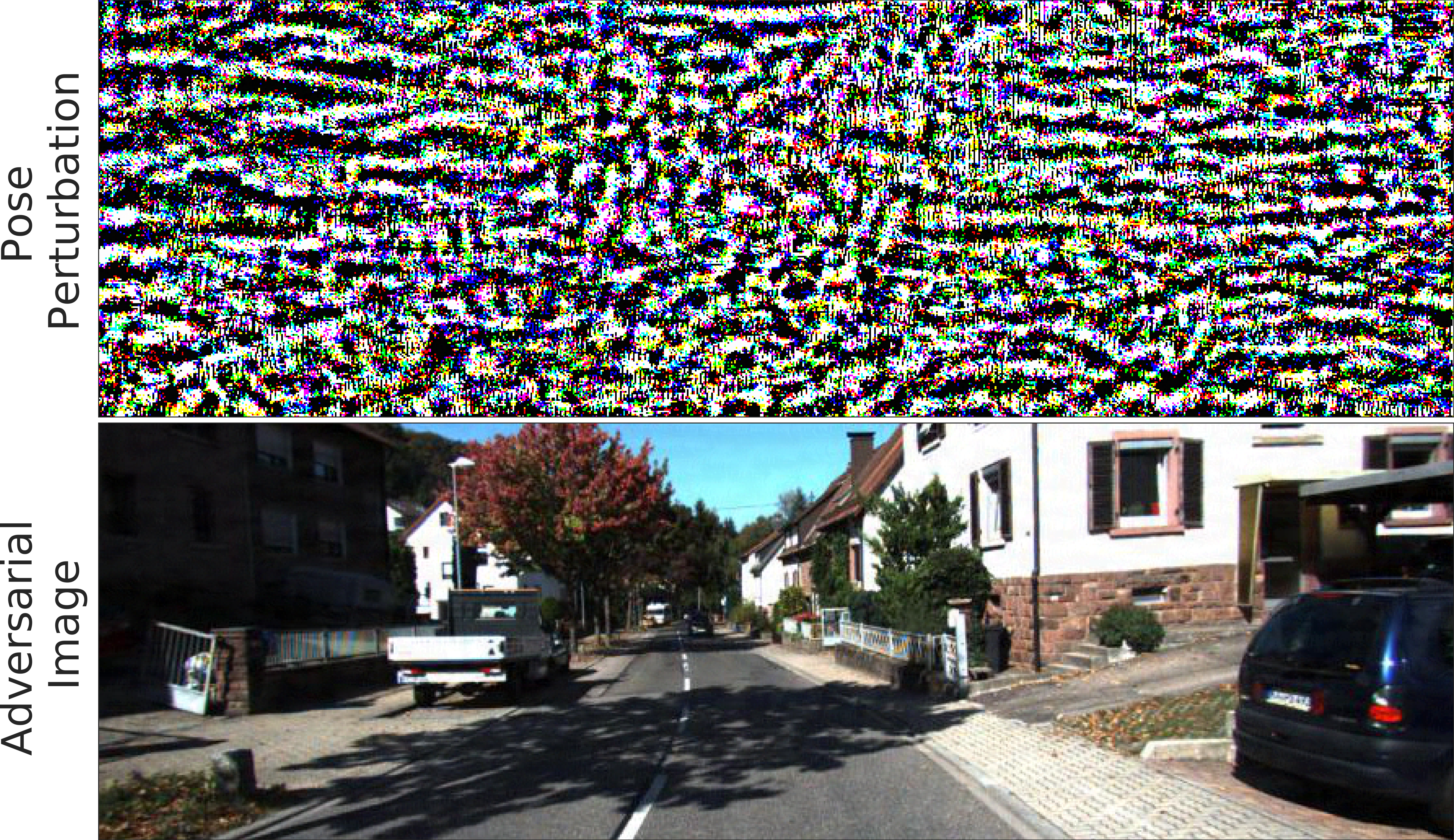} 
\end{subfigure}

\caption{Targeted attacks on depth and pose networks result in different types of adversarial perturbations. Example from KITTI odometry split at attack strength $\epsilon=4$. The attacks generate adversarial images with imperceptible changes.} 
\label{fig:perturbations}
\end{figure}

A growing number of works have studied this problem in the context of image classification~\cite{kurakin2017adversarial, yuan2019adversarial} and semantic segmentation~\cite{hendrik2017universal, arnab2018robustness}
proposing various attack as well as defense mechanisms~\cite{qin2019adversarial, tramer2019adversarial}.
However, these tasks are focused on predicting labels from a given set of categories.
A few recent works have also extended these attacks to the task of regressing pixel-wise monocular depth in supervised~\cite{zhang2020adversarial} as well as unsupervised settings~\cite{wong2020targeted}.
Unsupervised monocular depth estimation methods utilize image reconstruction error as form of supervision and hence train a camera pose estimation network simultaneously. These unsupervised depth and pose estimation networks have also been integrated with geometry-based visual odometry approaches for camera localization~\cite{li2019pose, bian2021unsupervised}. 
Recent works have investigated the effect of adversarial attacks against depth estimation~\cite{wong2020targeted, varma2022transformers}
Nevertheless, no study exists that investigates the effect of adversarial attacks against pose estimation networks. 

While pose estimation is also a regression task, it differs from depth estimation as it learns information about the camera by observing two overlapping images of a scene simultaneously. It outputs a rotation consisting of the roll, pitch, and yaw along with the translation in $x$, $y$, and $z$ from a pair of spatially overlapping consecutive input images, together forming a $4 \times 4$ homogeneous matrix representing the transformation from one image frame to another. This implies unique constraints on the range of output pose, different from depth estimation which regresses positive real values for each pixel in an input image. 
Hence, the type and impact of perturbations that adversarially attack pose estimation networks may be different from those that attack depth estimation networks.
Furthermore, while depth estimation networks make dense predictions for a single image, pose estimation networks make scene-level predictions from a pair of images.
Accordingly, any attack on the pose estimation network will impact the relative pose between images and subsequently the complete trajectory, unlike some attacks on depth estimation that only target specific objects or regions~\cite{wong2020targeted}. 

Thus, we study adversarial attacks on 
monocular pose estimation networks. We investigate the following questions:
\begin{itemize}
    \item Are pose estimation networks vulnerable to untargeted adversarial perturbations designed to cause mispredictions?
    \item Are pose estimation networks vulnerable to targeted adversarial perturbations designed to cause \textit{specific} mispredictions such as an incorrect direction of translation or rotation?
    \item Do targeted attacks on translation also impact the output rotation, and vice versa?
    \item Do adversarial images generated to fool pose estimation networks also fool simultaneously trained depth estimation networks, and vice-versa?
    \item Which of these adversarial attacks on pose estimation, if any, transfer to other pose estimation models with different architectures and losses?
\end{itemize}
Accordingly, we explore the use of Projected Gradient Descent (PGD)~\cite{DBLP:conf/iclr/MadryMSTV18} to fool networks into predicting an incorrect relative pose. 
Our experiments 
demonstrate that pose estimation networks are significantly vulnerable to untargeted as well as targeted adversarial attacks, as they can catastrophically alter the output trajectory. However, we note that the adversarial perturbations that attack depth are different from those that attack pose (see Figure~\ref{fig:perturbations}). Nonetheless, the attacks on pose estimation models are transferable to other models with different architecture and losses.
To the best of our knowledge, this study is the first to investigate adversarial attacks on pose estimation models.


\section{Related Work}

Several studies have explored the vulnerability of  deep neural networks to adversarial images that are imperceptibly different from the input in image classification~\cite{kurakin2017adversarial, yuan2019adversarial, DBLP:journals/corr/SzegedyZSBEGF13, DBLP:conf/iclr/MadryMSTV18, DBLP:journals/corr/GoodfellowSS14}.
This includes untargeted attacks that try to fool the networks into making any incorrect predictions, as well as targeted attacks that fool the networks into making specific incorrect predictions. 
Recently, these attacks have been extended to tasks such as semantic segmentation and object detection~\cite{hendrik2017universal,  xie2017adversarial, arnab2018robustness}.
Zhang \etal~\cite{zhang2020adversarial} further demonstrated a variety of untargeted, targeted, and universal attacks for monocular depth estimation. Wong \etal~\cite{wong2020targeted} proposed multiple targeted attacks for unsupervised depth estimation, and uncovered biases learned by unsupervised depth estimation methods. However, none of these works demonstrate adversarial attacks on monocular pose estimation. Thus, we formulate and analyze adversarial attacks for unsupervised pose estimation methods.

\section{Methodology}
We first briefly discuss the training of unsupervised depth and pose estimation networks.
Thereafter, we formulate the adversarial attacks on pose estimation and present our methods for untargeted and targeted attacks, adapted from the methods on classification and depth estimation. 

\subsubsection{Unsupervised monocular depth and pose estimation}
\label{sec:unsupervised_method}
Unsupervised monocular pose estimation is learned in parallel with monocular depth estimation. A training sample for the depth- and pose-estimation networks consists of  consecutive RGB image frames $\{I_a, I_b\}$ from a video sequence. The pose estimation network $f_P$ learns to estimate the relative rotation and translation between a pair of consecutive images. The rotation $R$ consisting of the roll $\phi$, pitch $\theta$, and yaw $\psi$ along with the translation $t$ in $x$, $y$, and $z$, form the relative transformation matrix $T$. The depth estimation network $f_D$ learns to estimate pixel-wise depth for a single input image. The depth and pose estimation networks are connected together via the perspective projection transform that is used for synthesizing a target view from the nearby source view~\cite{bian2021unsupervised}. 

Photometric loss between the synthesized and original target view~\cite{bian2021unsupervised}, $L_{p}$ is used to train the depth and pose networks together. Additionally, an edge-aware smoothness loss~\cite{zhou2017unsupervised} $L_{s}$ is applied to regularize the estimated depth. Finally, a geometric consistency loss $L_{g}$ is used by some methods~\cite{bian2021unsupervised} to improve the consistency of estimates across the video sequence. Thus, the total loss used to train the depth and pose networks is given by:
\begin{equation}
\label{eq:loss_networks}
L_{\text{train}} = w_1 \cdot L_{p} + w_2 \cdot L_{s} + w_3 \cdot L_{g}, 
\end{equation}
where $w_1$, $w_2$, and $w_3$ are the hyper-parameters. 

 
 \subsection{Adversarial attacks formulation}
 Given a pre-trained pose estimation network, $f_P: \mathbb{R}^{H \times W \times 6} \rightarrow  \mathbb{R}^3 \bigtimes [-\pi, \pi)^3$, the objective is to find a perturbation $\xi(x) \in \mathbb{R}^{H \times W \times 6}$, as a function of the input image pair $x$, such that $x^{\text{adv}} = x + \xi(x)$, which changes the pose prediction to an incorrect output, $T(x^{\text{adv}}) \neq T(x)$. This is accomplished 
 by maximizing an adversarial loss function $L^{\text{adv}}$. 
 To keep the adversarial changes imperceptible, the perturbation is bound such that $\left\lVert \xi \right\rVert_{\infty} < \epsilon$, and the pixel values are in valid range $[0, 255]$.
 
 We use Projected Gradient Descent~\cite{DBLP:conf/iclr/MadryMSTV18} at multiple attack strengths.
 Following ~\cite{kurakin2017adversarial}, the perturbation is accumulated over $min(\epsilon + 4, \lceil1.25\cdot\epsilon\rceil)$ iterations with a step-size of $1$.  Accordingly, the adversarial image is initialized as $x^{adv}_0 = x$, and is updated at each iteration 
 \begin{equation}
 \label{eq:ifgsm}
x^{\text{adv}}_{i+1} = \text{CLIP}(x_i^{\text{adv}} + \alpha \cdot \text{sign}(\nabla L^{\text{adv}}(x_{\text{adv}}, f_P, f_D)), \epsilon)
\end{equation}
where $\text{CLIP}(x, \epsilon)$ limits $x$ within $[x -\epsilon, x +\epsilon]$, and $L^{\text{adv}}$ is the adversarial loss function, formulated differently for each attack  as described below.
 
\subsection{Untargeted Adversarial Perturbations}
For the untargeted attack, the objective is to generate adversarial images that cause errors on the test set. 
To do so, we utilize the reprojection loss used to train the networks as the adversarial loss function for generating the adversarial images. Hence, 
\begin{equation}
\label{eq:untargeted_loss}
L(x^{\text{adv}}, f_D, f_P) = L^{\text{train}}
\end{equation}
Designed for an unsupervised learning paradigm, ground truth depths or poses are not used during generation of the adversarial images. 



\subsection{Targeted Adversarial Perturbations}
Misjudging the pose of the vehicle can prove fatal in an autonomous navigation scenario. The objective of targeted attacks is to fool the network into predicting a \textit{specific} incorrect pose. 
We consider three kinds of pose targets $T^{\text{tgt}}(x)$, formed by suitably altering the predictions of the trained network on the clean inputs. 
First, we want to force the network to be fooled into predicting vehicle yaw in the opposite direction. Second, we want to fool the network into predicting that the vehicle is moving backwards. Third, we want to fool the network into predicting inverted pose transformation i.e complete reverse translation as well as rotation along all axes.

The relative transformation, $T^{\text{rel}}$ between the original prediction $T(x)$ and the target pose prediction $T^{\text{tgt}(x)}$ is computed as:
\begin{equation}
\label{eq:transformation_matrix} 
T^{\text{rel}}(x) = T^{-1}(x) T^{\text{tgt}}(x) = 
 \begin{bmatrix}
R^{rel}(x) & t^{rel}(x) \\
\mathbf{0} & 1
\end{bmatrix}
\end{equation}
The loss $L(x^{\text{adv}}, T^{\text{rel}}(x), f_P)$ for the targeted attacks is then computed accordingly. Similar to the untargeted attack, we formulate the loss function for the unsupervised learning paradigm without taking ground truth pose as input. 

\subsubsection{Inverting the yaw}
The most prominent rotation for a driving scenario is the \textit{yaw}, when the vehicle makes a turn or changes lanes. We examine the possibility of fooling the network into predicting the opposite yaw to what was originally output. The target yaw is defined as:
\begin{equation}
\label{eq:fn_invert_yaw}    
\psi^{\text{tgt}}(x) = \texttt{invert\_yaw}(f_P(x)) = - \psi
\end{equation}

The target yaw along with the originally predicted roll, pitch, and translations are then converted to the target relative transformation. 
Accordingly,  the loss used for this attack is given by:
\begin{equation}
\label{eq:loss_invert_yaw}
L^{\text{adv}} = r_{\text{err}}(T^{\text{rel}}) = \cos^{-1}\left(\dfrac{\texttt{trace}(R^{\text{rel}}) - 1}{2}\right).
\end{equation}


\subsubsection{Moving backwards}
The most prominent translation for a driving scenario is the linear forward motion. 
We examine the possibility of fooling the network into predicting motion in the backwards direction. The target translation is defined as 
\begin{equation}
\label{eq:fn_move_backwards}
{t_z}^{\text{tgt}}(x) = \texttt{move\_backwards}(f_P(x)) = - t_z
\end{equation}
The target translation in $z$ along with the originally predicted $t_x, t_y$ and rotations are then converted to the target relative transformation. 
Accordingly, the loss used for this attack is given by:
\begin{equation}
\label{eq:loss_move_backwards}
L^{\text{adv}} =  t_{\text{err}}(T^{\text{rel}}) 
= \sqrt{(t^{\text{rel}}_x)^2 + (t^{\text{rel}}_y)^2 + (t^{\text{rel}}_z)^2}
\end{equation}


\subsubsection{Inverting the pose}
Since the pose network output is dependent upon the order of the input images, we examine the possibility of predicting the inverse relative pose transformation between a pair of images. 
Therefore the target is given by:
\begin{equation}
\label{eq:fn_invert_pose}
T^{\text{tgt}}(x) = \texttt{invert}(f_P(x)) = T^{-1}(x)
\end{equation}
Then, the loss used for this attack is given by:
\begin{equation}
\label{eq:loss_invert_pose}
L^{\text{adv}} =  t_{\text{err}}(T^{\text{rel}}) + r_{\text{err}}(T^{\text{rel}})
\end{equation}

\section{Results}
\subsection{Experimental Setup}
We evaluate adversarial attacks on the KITTI odometry split~\cite{Geiger2012CVPR}. This dataset consists of images captured in a driving scenario across the city as well as highways.  The KITTI odometry sequences $00$ to $08$ are used for training, while sequences $09$ and $10$ are used for evaluation. The evaluation sequences have a total of $1591$ and $1201$ test images respectively. We utilize SC-Depth~\cite{bian2021unsupervised} models pre-trained\footnote{\url{https://github.com/JiawangBian/SC-SfMLearner-Release\#pretrained-models}} on image size of $832\times 252$ for generating adversarial perturbations. The simultaneously trained depth and pose networks use ResNet50~\cite{he2016deep} encoders 
with photometric, depth smoothness, and geometry consistency losses. 
The experiments are performed on SC-Depth as it is one of the state-of-the-art methods and it utilizes all the common training losses used by other methods. 

\subsection{Untargeted Attack}
\begin{figure}[t]
  \centering
  \begin{tabular}{@{}c@{}}
    \includegraphics[width=0.83\linewidth]{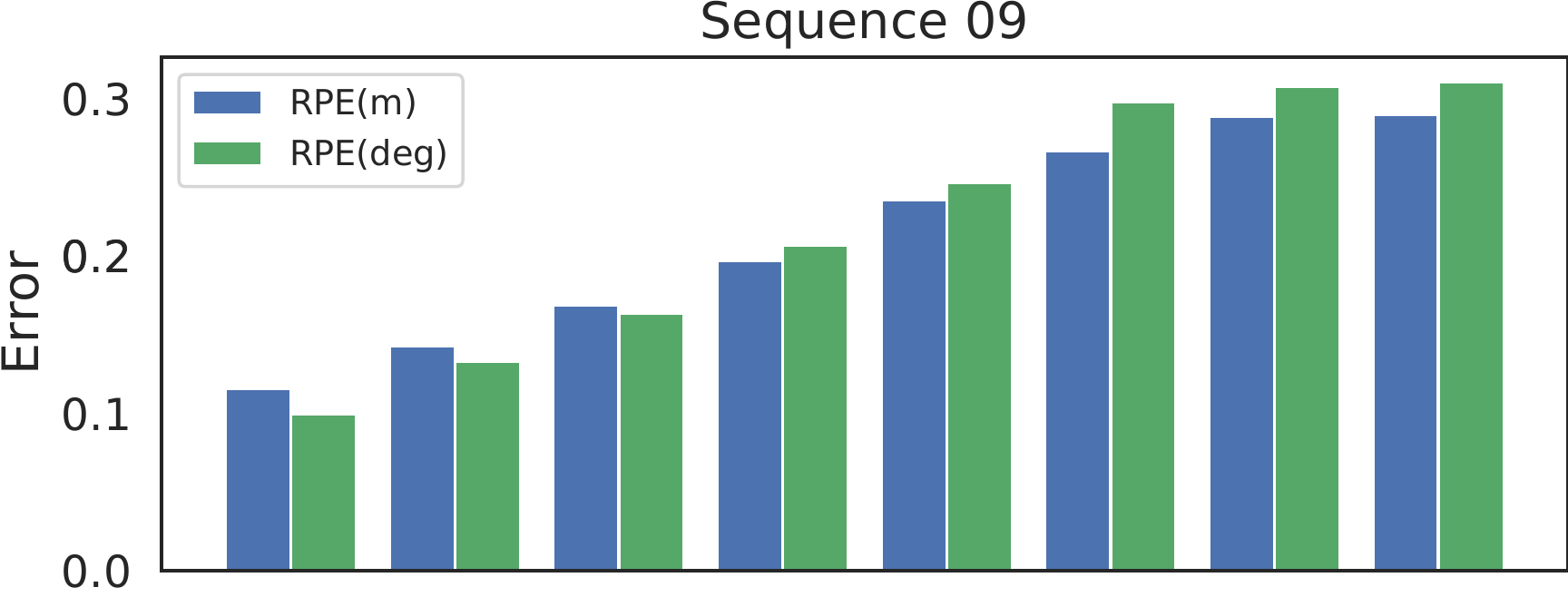} 
  \end{tabular}


  \begin{tabular}{@{}c@{}}
    \includegraphics[width=0.83\linewidth]{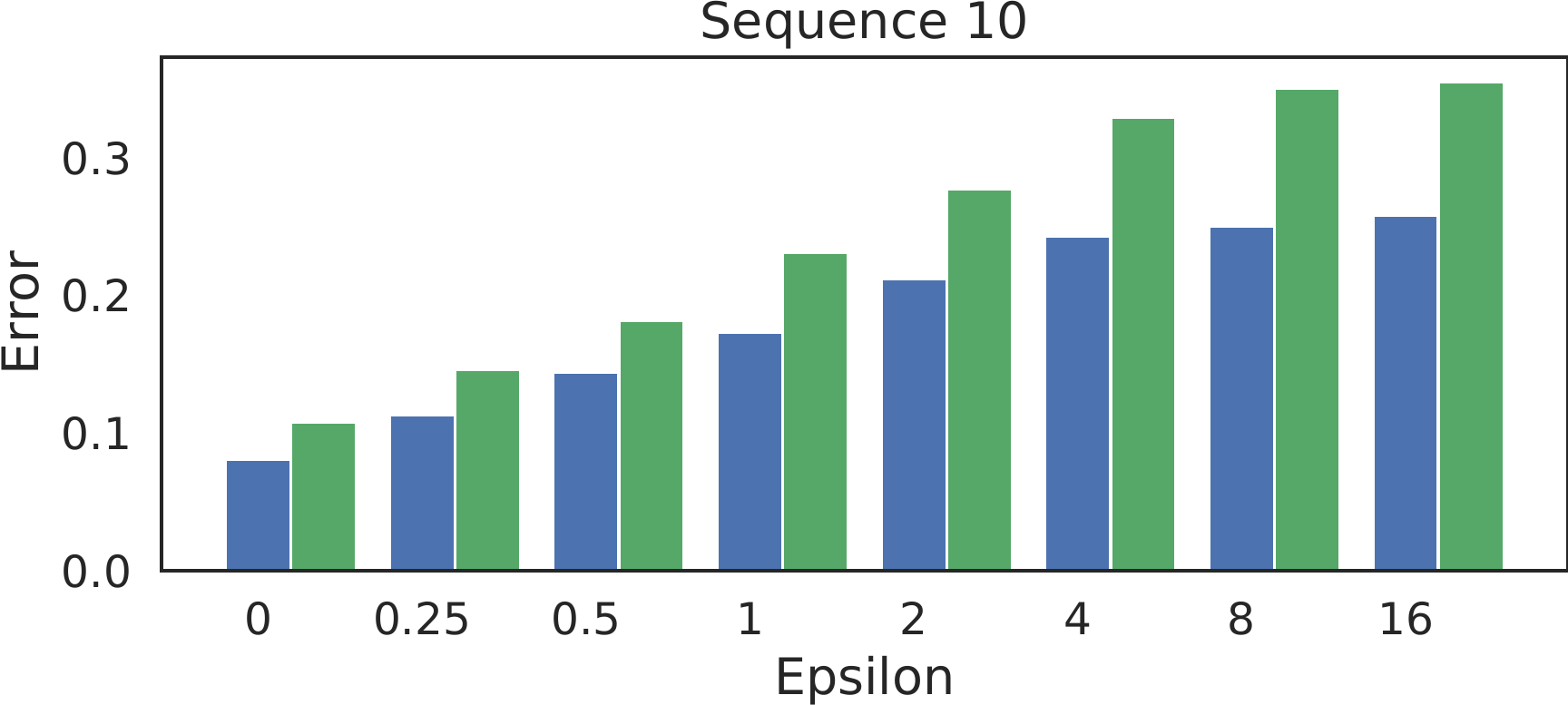} 
  \end{tabular}

  \caption{Relative Pose Error (RPE) for untargeted attack on pose network. Evaluated on  KITTI
odometry split.Higher value indicates higher impact of the attack.}
  \label{fig:pgd_vo}
\end{figure}


In this section, we examine  if pose estimation networks are vulnerable to untargeted adversarial perturbations.
We perform the untargeted attack for $\epsilon \in \{0.25, 0.5, 1, 2, 4, 8, 16\}$.
Figure \ref{fig:pgd_vo} shows the average Relative Pose Error (RPE)~\cite{sturm2012evaluating} for translation (m) and rotation (deg) corresponding to different attack strengths for both test sequences. 
We observe that the errors increase with the attack strength. The translation error at $\epsilon=16$ becomes $2.5\times$ and $3.19 \times$ for Sequences 09 and 10 respectively. The rotation error similarly increases to $3.11 \times$ and $3.29\times$ for Sequences 09 and 10 respectively.  

This confirms that the pose network is vulnerable to an untargeted attack.

\subsection{Targeted Attacks}

In this section, we examine if pose estimation networks are vulnerable to targeted adversarial perturbations that fool the network into predicting a specific output relative pose. We also examine if targeted attacks on translation impact the output
rotation, and vice versa. 
Targeted attacks further allow to study the potential biases learned by the network~\cite{wong2020targeted}. We perform targeted attacks for $\epsilon=\{ 1, 2, 4\}$. 

Table~\ref{tabl:targeted_vo} shows the impact of targeted attacks through the 
ratio of RPE on adversarial images and the RPE on clean images ($\epsilon=0$).
Higher ratio indicates a higher vulnerability to the attack. The trajectories generated from the pose predictions on adversarial images are shown in Figure~\ref{fig:targeted_vo_trajectories}.

\subsubsection{Invert Yaw}
We observe that the `Invert Yaw' targeted attack increases the average errors for all attack strengths. For Sequence 09, the average RPE(deg) increases to $18.46\times$, while the average RPE(m) increases to $4.83\times$. Similarly, for Sequence 10, the average RPE(deg) increases to $17.83\times$, while the average RPE(m) increases to $5.75 \times$. This shows that the targeted attack on rotation catastrophically alters the pose output. It also shows that the translation error is impacted by attack on rotation.
It can be inferred that the pose networks learns features that relate translation to the rotation between the image pair.
Figure~\ref{fig:targeted_vo_trajectories:invert_yaw} demonstrates the qualitative impact of `Invert Yaw' targeted adversarial attack. The attack is effective in altering the global consistency (i.e. shape) of the trajectory even at $\epsilon=1$, introducing loops in its output. Attacks at $\epsilon=\{2, 4\}$ make the output completely different from the expected trajectory.  

\begin{table}[t]
\centering
\resizebox{\linewidth}{!}{
\begin{tabular}{|c|c|c|c|c|c|}
\cline{3-6}
\multicolumn{2}{c|}{\textbf{}} & \multicolumn{4}{c|}{\textbf{Ratio}} \\
\cline{3-6}
\multicolumn{2}{c|}{\textbf{}} & \multicolumn{2}{c|}{\textbf{Sequence 09}} & \multicolumn{2}{c|}{\textbf{Sequence 10}} \\ \hline
\textbf{Attack} & \textbf{$\epsilon$} & RPE (m) & RPE (deg) & RPE (m) & RPE (deg) \\ \hline \hline
\multirow{3}{*}{\textit{Invert Yaw}} & 1 & 1.22 & 5.07 & 2.36 & 4.74 \\
 & 2 & 3.73 & 9.72 & 3.51 & 9.37 \\
 & 4 & 4.83 & 18.46 & 5.75 & 17.83 \\ \hline
\multirow{3}{*}{\textit{Move Backwards}} & 1 & 1.07 & 1.36 & 1.23 & 1.30 \\
 & 2 & 1.16 & 2.12 & 1.62 & 1.94 \\
 & 4 & 2.16 & 3.65 & 2.30 & 3.16 \\ \hline
\multirow{3}{*}{\textit{Invert Pose}} & 1 & 1.24 & 2.56 & \multicolumn{1}{c|}{1.44} & 2.29 \\
 & 2 & 1.20 & 4.63 & 2.17 & 4.18 \\
 & 4 & 1.51 & 8.35 & 3.07 & 7.75 \\ \hline
\end{tabular}
}
\caption{Ratio of Relative Pose Error (RPE) for targeted attacks on pose network. Evaluated on KITTI
odometry split. Higher value indicates higher impact of the attack. }
\label{tabl:targeted_vo}
\end{table}

\subsubsection{Move Backwards}
We further observe that the `Move Backwards' targeted attack similarly increases the average errors for all attack strengths. For Sequence 09, the average RPE(m) increases to $2.16\times$, while the average RPE(deg) increases up to $3.65\times$. Similarly for Sequence 10, the average RPE(m) increases up to $2.3\times$, while the average rotation error increases up to $3.16\times$. This shows that the targeted attack on translation catastrophically alters the pose output. It shows that the rotation error is also impacted by attack on translation, just as the translation was impacted by the attack on rotation.
This implies that the features learned by the pose network relate rotation and translation with each other. 
However, note that the error caused by attacking translation is less than that caused by attacking yaw. Forcing the network to generate predictions that correspond to the vehicle moving backwards is more difficult than forcing it to generate predictions corresponding to turning in the opposite direction. 

Figure~\ref{fig:targeted_vo_trajectories:move_backwards} additionally shows qualitative impact of `Move Backwards' targeted adversarial attack. The attack is not as effective in altering the shape of the trajectory at $\epsilon=1$. 
However, both the trajectory drift and global consistency are strongly altered at higher attack strengths. 

\subsubsection{Invert Pose}
\begin{figure*}[htbp]
\centering
\raisebox{-0.025\linewidth}{\rotatebox{90}{\small{Sequence 09}}}
\begin{subfigure}{.3\textwidth}
  \centering
  \includegraphics[width=0.85\linewidth]{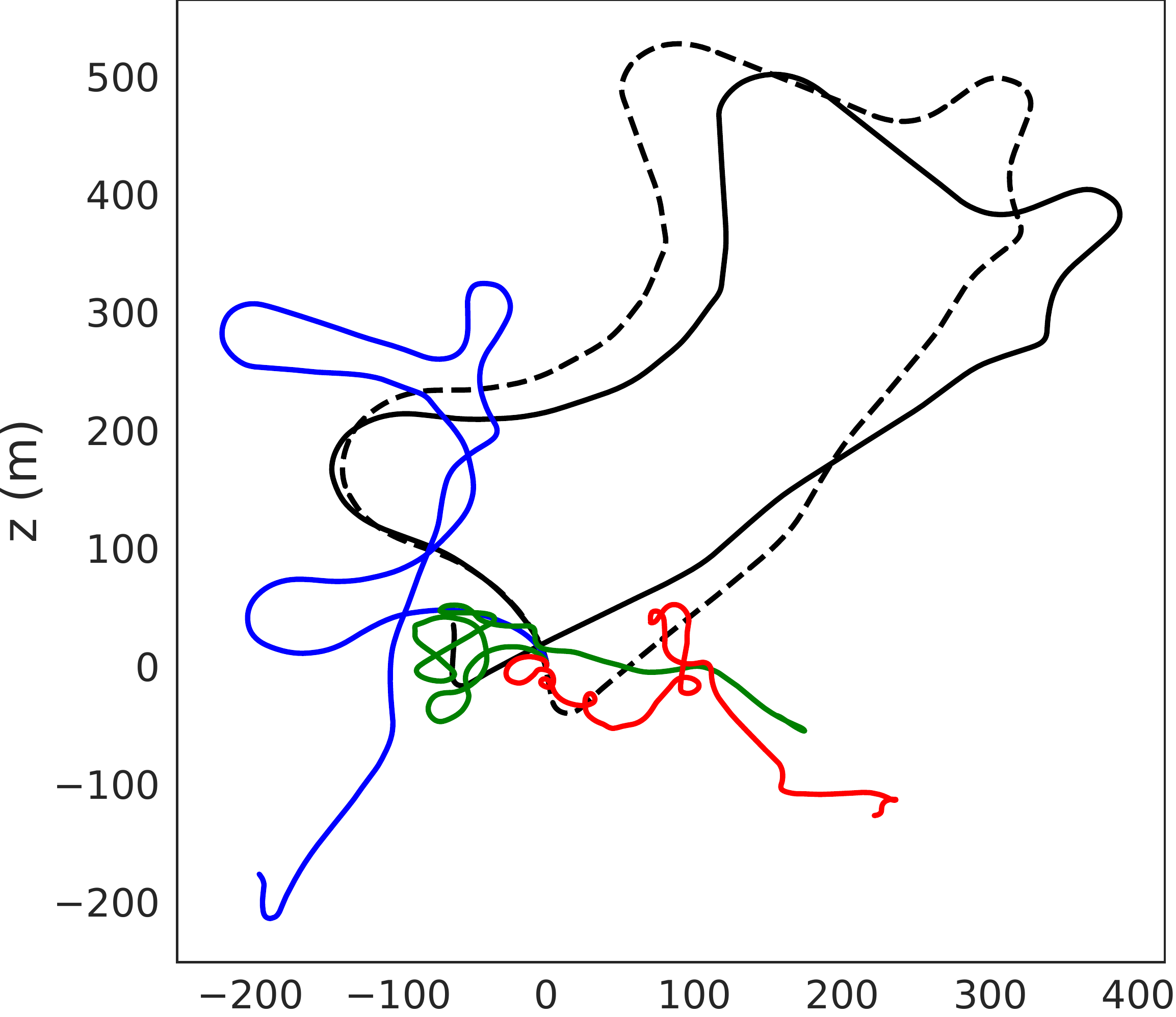}  
\end{subfigure}
\begin{subfigure}{.3\textwidth}
  \centering
  \includegraphics[width=0.85\linewidth]{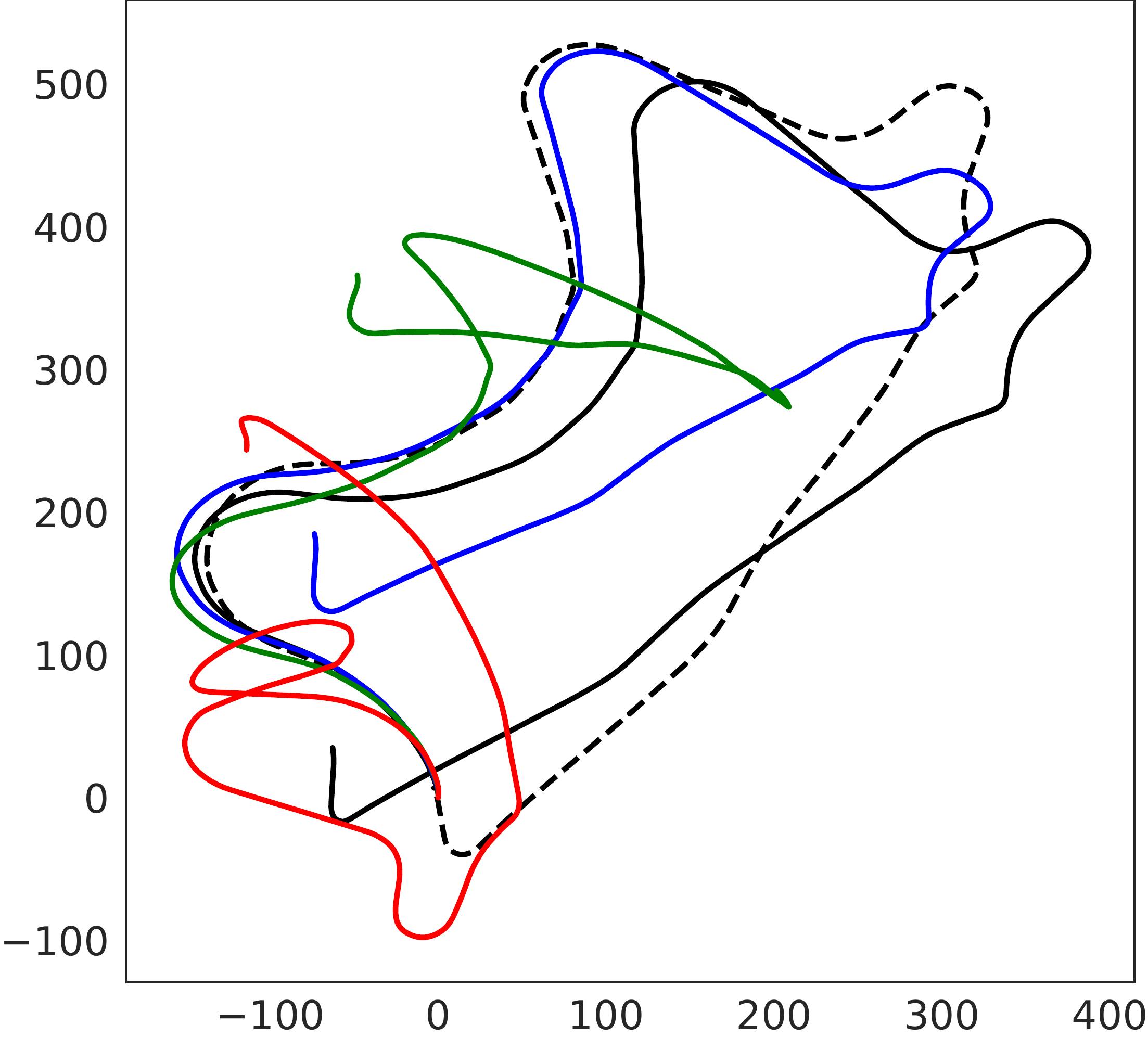}  
\end{subfigure}
\begin{subfigure}{.3\textwidth}
  \centering
  \includegraphics[width=0.85\linewidth]{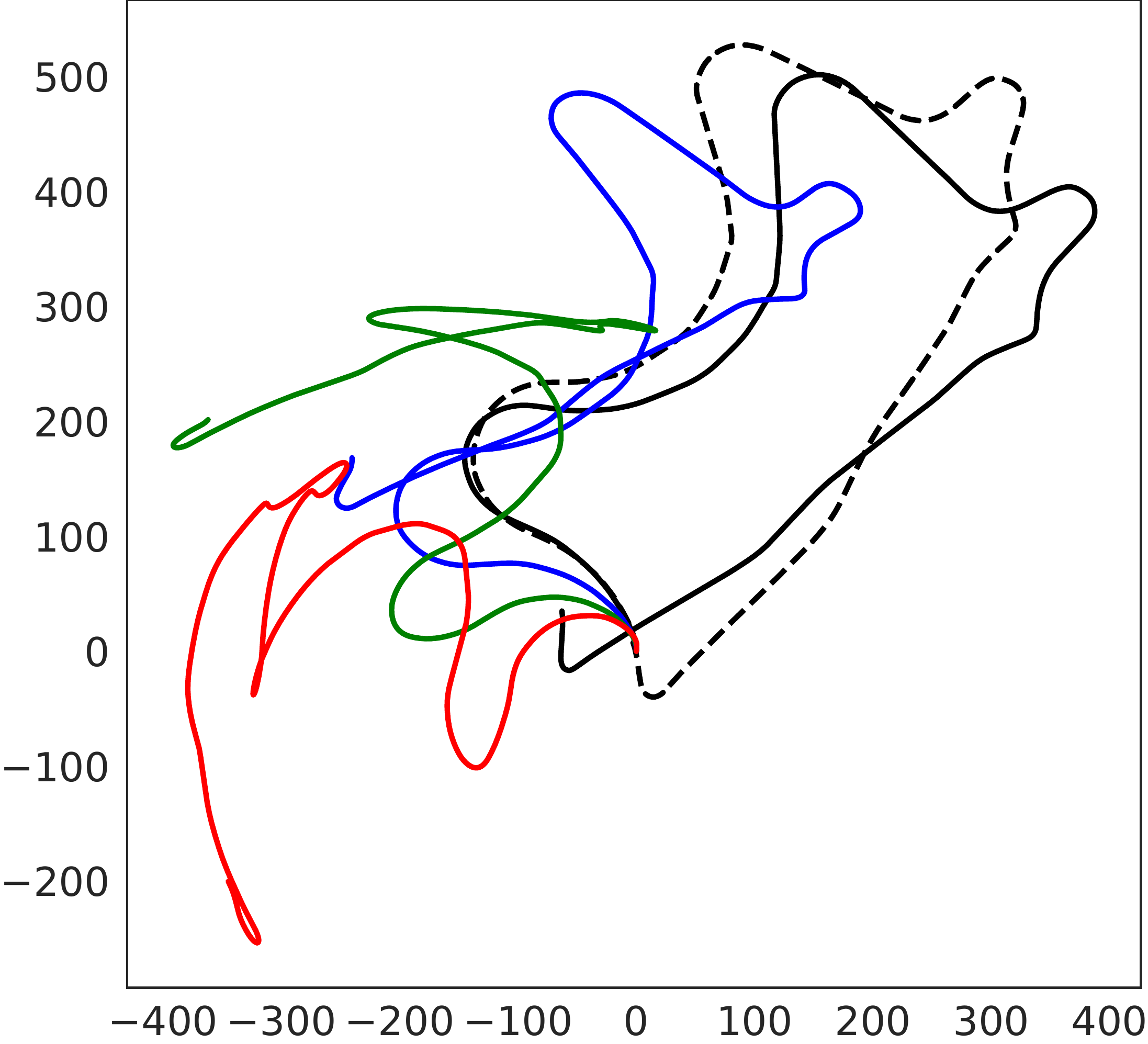}  
\end{subfigure}

\medskip

\raisebox{-0.025\linewidth}{\rotatebox{90}{\small{Sequence 10}}}
\begin{subfigure}{.3\textwidth}
  \centering
  \includegraphics[width=0.85\linewidth]{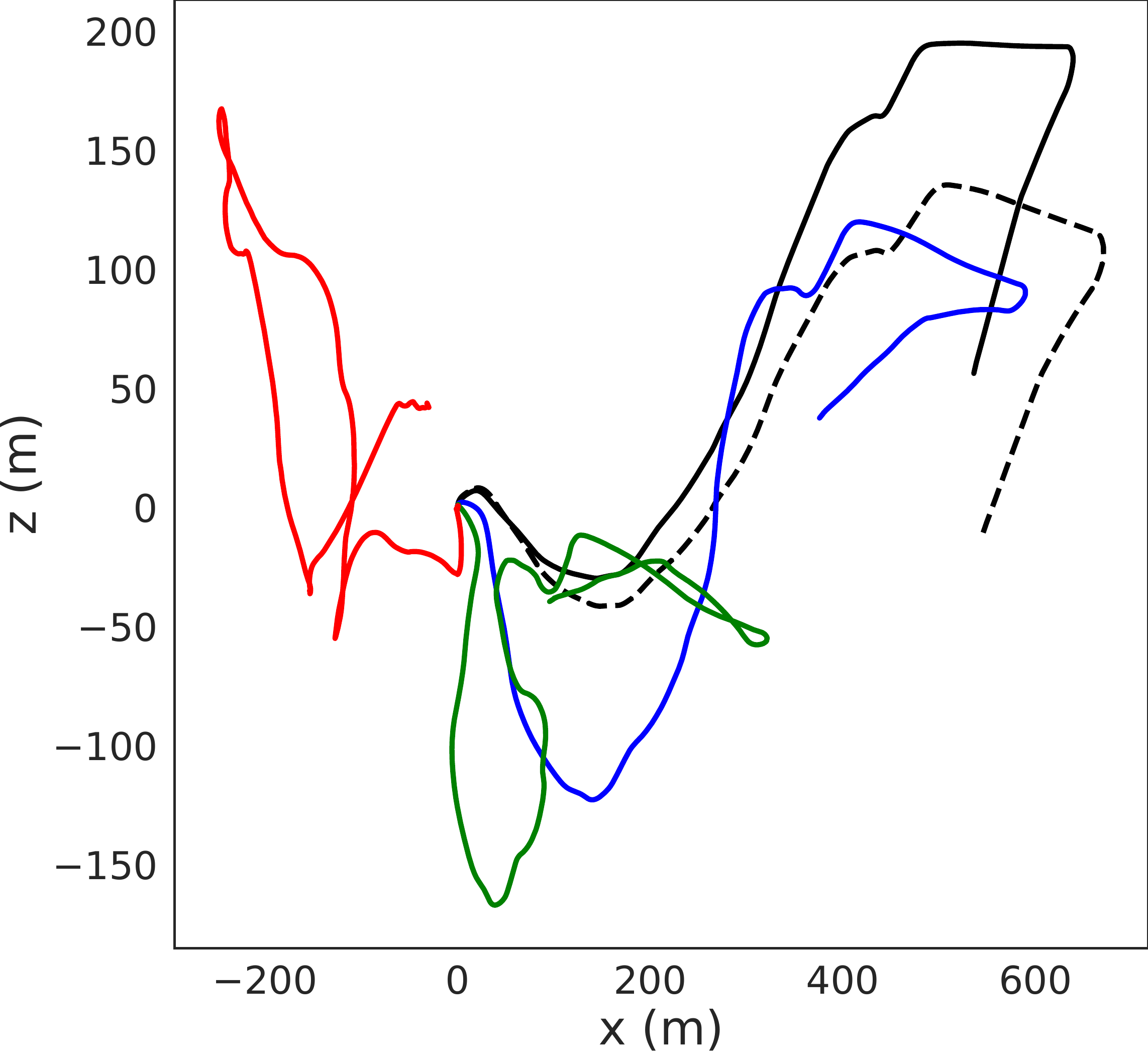}  
  \caption{Invert yaw}
  \label{fig:targeted_vo_trajectories:invert_yaw}
\end{subfigure}
\begin{subfigure}{.3\textwidth}
  \centering
  \includegraphics[width=0.85\linewidth]{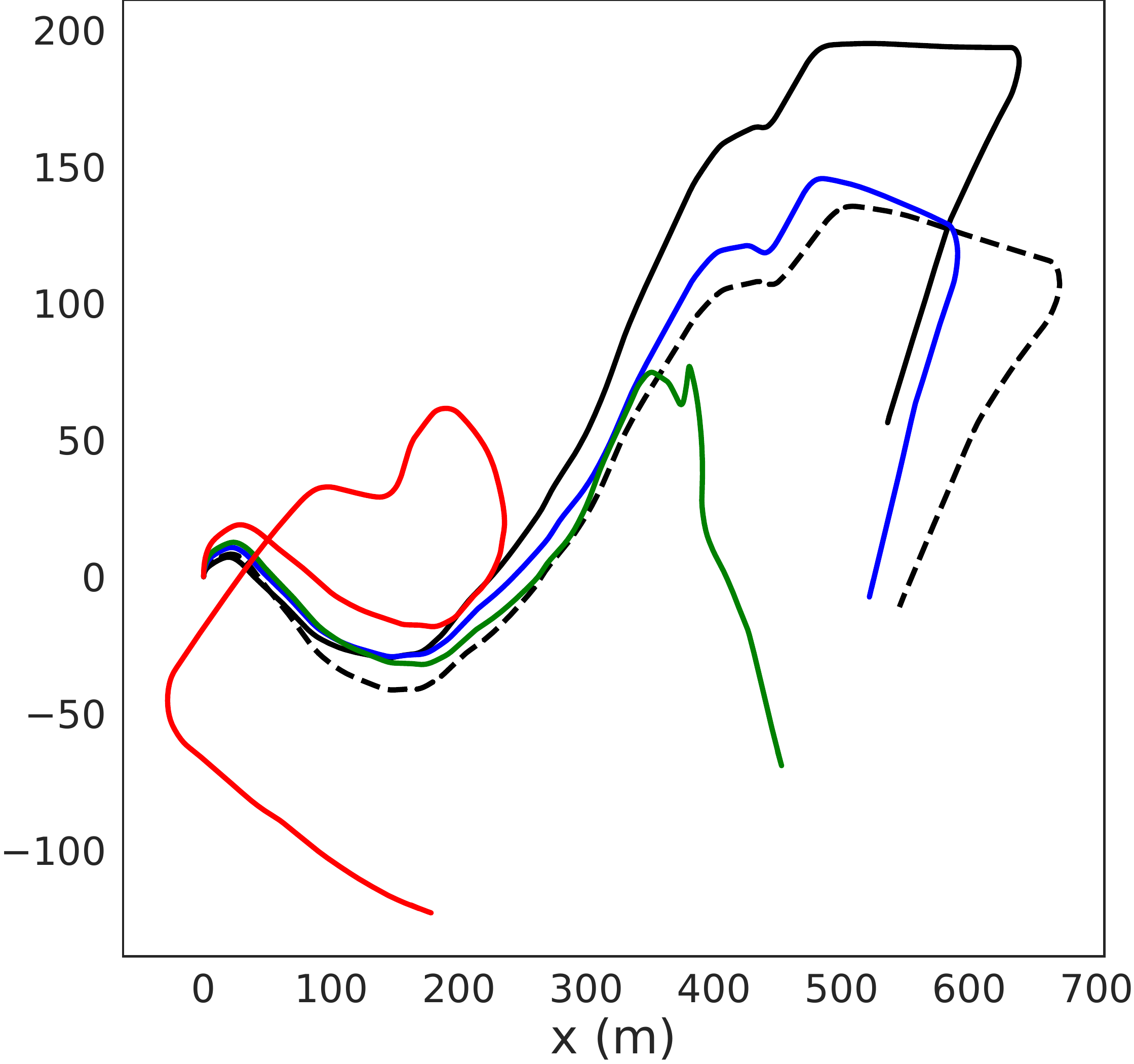}  
  \caption{Move backwards}
  \label{fig:targeted_vo_trajectories:move_backwards}
\end{subfigure}
\begin{subfigure}{.3\textwidth}
  \centering
  \includegraphics[width=0.85\linewidth]{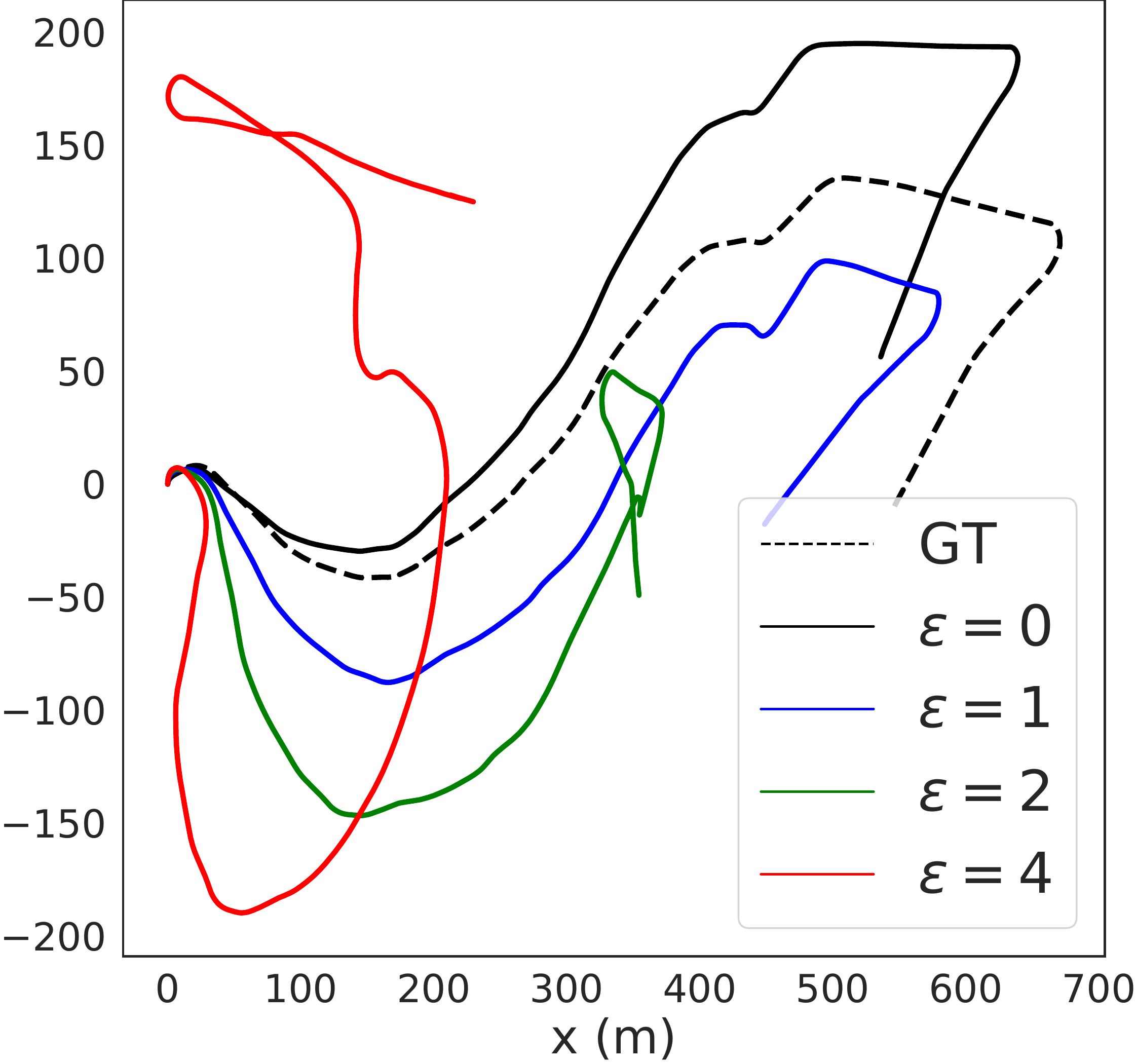}  
  \caption{Invert pose}
  \label{fig:targeted_vo_trajectories:invert_pose}
\end{subfigure}

\caption{ Origin-aligned trajectories computed for targeted attacks on pose network. Evaluated on KITTI
odometry split.}
\label{fig:targeted_vo_trajectories}
\end{figure*}

We further observe that the `Invert Pose' targeted attack also increases the average errors for all attack strengths. For Sequence 09, the average RPE(m) increases to $1.51\times$, while the average RPE(m) increases much more up to $8.35\times$. Similarly for Sequence 10, the average RPE(m) increases to $3.07\times$, while the average RPE(m) increases to $7.75\times$.  This shows that the targeted attack on the complete pose also catastrophically alters the pose output. The rotation is affected more than the translation, as was also the case for the previous attacks. 
Figure~\ref{fig:targeted_vo_trajectories:invert_pose} additionally shows qualitative impact of `Invert Pose' targeted adversarial attack. We observe at $\epsilon=1$ that this attack is more effective than `Move Backwards' attack in altering the global consistency, but not as effective as the `Invert Yaw' attack. 
 

We conclude that targeted attacks on translation and rotation both lead to notable errors in the predictions. We also note that targeted attacks are more severe than untargeted attacks. Different types of targeted attacks alter the resultant trajectories in different ways as depicted in Figure~\ref{fig:targeted_vo_trajectories}.
Furthermore, the features learnt by the pose network relate rotation and translation with each other. Hence, attacking translation results in degradation of predicted rotation and vice versa. 

\subsection{Cross-task Attacks}
In this section, we additionally examine if adversarial images generated to fool pose estimation networks also fool depth estimation networks, and vice versa. This is done by evaluating the depth network on adversarial images generated through targeted attacks on the pose network, and vice versa. We perform these cross-task targeted attacks at $\epsilon=4$. 
Recall that the untargeted attack uses the training loss as adversarial loss, and is same on the depth and pose networks (see Eq.~\ref{eq:untargeted_loss}). Hence, we do not consider it for cross-task attacks.

\subsubsection{Attacking Pose, Evaluating Depth}
Generation of the adversarial examples from targeted attacks on pose network requires image pairs as input.  However, the depth is estimated for each image individually. 
We compute the depth corresponding to the first adversarial image for all pairs of the test sequence. For the last pair, we additionally compute the depth corresponding to the second adversarial image.
giving a total of $n$ depth maps for the test sequence.  
The ratio of root mean squared errors (RMSE)~\cite{zhou2017unsupervised} on the adversarial and the clean images is used to evaluate this cross-task attack.
As shown in Table~\ref{tab:cross_task} the maximum ratio of RMSEs is $1.04$ indicating that the depth network is not vulnerable to the adversarially perturbed images generated to fool the pose network. 

\subsubsection{Attacking Depth, Evaluating Pose}
We also perform targeted attack for the depth network and generate adversarial images. Following~\cite{varma2022transformers}, we use horizontally and vertically flipped predictions as targets to generate the adversarial images. Thereafter, the generated images are passed through the pose prediction network in a sequential order as image pairs to compute the output trajectory. 
As earlier, the ratios of RPE (m) and RPE (deg) on the adversarial and the clean images are used to evaluate this cross-task attack. As shown in Table~\ref{tab:cross_task}, the maximum ratio of RPE (m) is $1.04 $ and the maximum ratio of RPE (deg) is $1.02 $, indicating that the pose network is also not vulnerable to attack by the adversarially perurbed images generated to fool the depth network. 

\begin{table}[t!h]
\centering
\resizebox{\linewidth}{!}{
\begin{tabular}{|c|c|c|c|c|c|}
\hline
\textbf{Attack} & \textbf{Eval} & \textbf{Target} & \textbf{Ratio} & \textbf{Sq 09} & \textbf{Sq 10} \\ \hline \hline
\multirow{3}{*}{Pose} & \multirow{3}{*}{Depth} & \textit{Invert Yaw} & \multirow{3}{*}{RMSE} & 1.03 & 1.03 \\ \cline{3-3} \cline{5-6} 
 & & \textit{Move Backwards} & & 1.04 & 1.03 \\ \cline{3-3} \cline{5-6}  
 & & \textit{Invert Pose} & & 1.03 & 1.03 \\ \hline
\multirow{4}{*}{Depth} & \multirow{4}{*}{Pose} & \multirow{2}{*}{\textit{Flip Horizontal}} & RPE(m) & 1.04 & 1.02                 \\\cline{4-6} 
 & & & RPE(deg) & 1.02 & 1.01 \\ \cline{3-6} 
 & & \multirow{2}{*}{\textit{Flip Vertical}} & RPE(m) & 1.02 & 1.02 \\\cline{4-6} 
 & & & RPE(deg) & 1.01 & 1.00 \\ \hline
\end{tabular}
}
\caption{Ratios of Relative Pose Error (RPE) for cross-task attacks on depth and pose networks. Evaluated at attack strength $\epsilon=4$ on KITTI
odometry split. Higher value indicates higher impact of the attack. }
\label{tab:cross_task}
\end{table}

Figure~\ref{fig:perturbations} also illustrates how the adversarial perturbations generated to attack depth and pose estimation networks are different, which might indicate why they do not succeed in attacking the other network. This implies that access to one of the depth or pose models (trained simultaneously in an unsupervised manner), does not threaten the security of the other. 

\begin{figure*}[t]
\centering
\includegraphics[width=.9\textwidth]{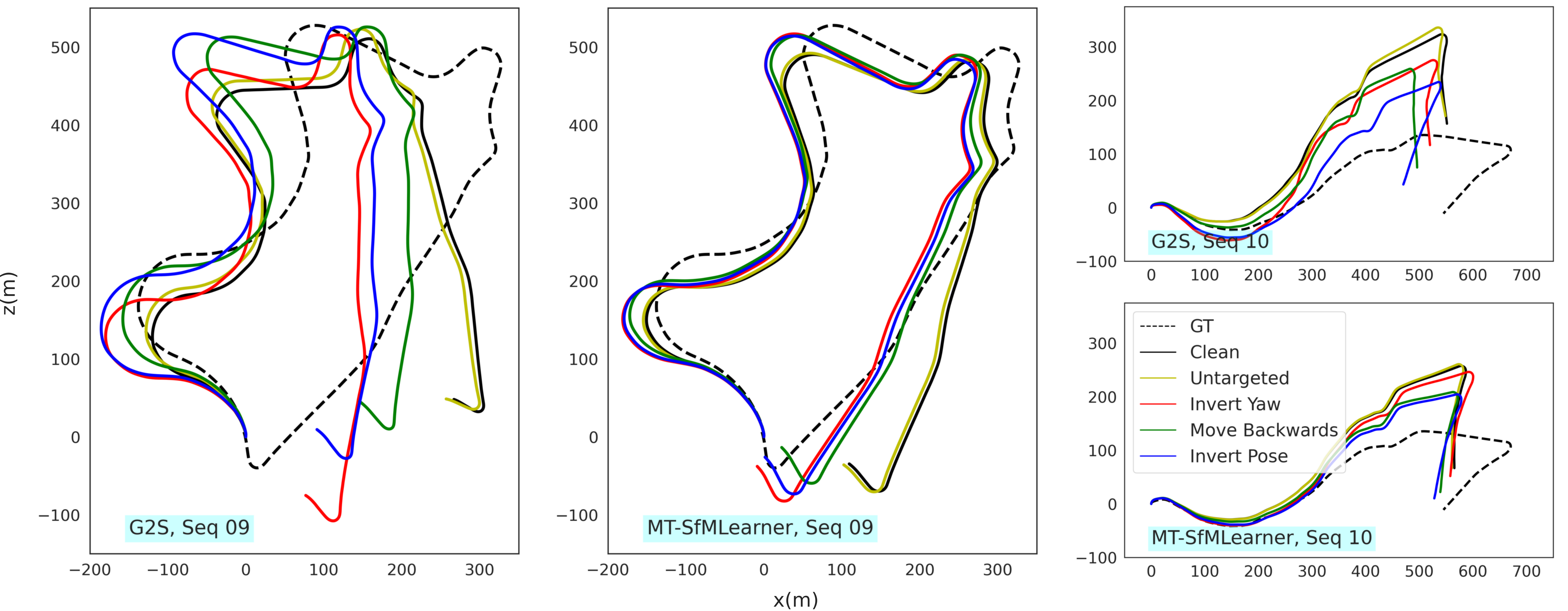}  
\caption{Transferability of adversarial attacks across pose networks. Origin-aligned trajectories for G2S and MT-SfMLearner evaluated on adversarial images  from SC-Depth. Evaluated at attack strength $\epsilon=4$ on KITTI
odometry split.}
\label{fig:transferability}
\end{figure*}

\subsection{Transferability}
In this section, we examine if the adversarial attacks on pose estimation transfer to other pose estimation models with different architectures and losses. We consider two additional pose-estimation models, namely, G2S~\cite{chawla2021multimodal} and MT-SfMLearner~\cite{varma2022transformers}.

G2S uses a similar CNN-based architecture as SC-Depth. However, it does not have a geometric consistency loss. Instead, it is trained using a multimodal GPS-to-Scale loss for introducing scale-consistency and -awareness in the predictions. MT-SfMLearner uses Transformer-based
architecture which is a heavier model. 
MT-SfMLearner neither uses the geometric consistency loss as SC-Depth, nor the GPS-to-Scale loss. 

For examining the transferability of the pose adversarial perturbations, we pass the adversarial images generated from SC-Depth model to infer on G2S and MT-SfMLearner models.
Figure~\ref{fig:transferability} shows the trajectories computed on adversarial images generated by attacking SC-Depth at $\epsilon=4$.

We observe that untargeted adversarial perturbations do not transfer to other models. However, targeted adversarial perturbations, which were found to be stronger for SC-Depth, transfer to other models with different architectures and training losses. They result in an increased odometry drift, but do not catastrophically alter the shape of the trajectories. 

\section{Conclusion}
\label{sec:conclusion}
We investigate adversarial attacks on pose estimation models. 
Via our experiments on the KITTI odometry split, we demonstrate that these models can be attacked using untargeted perturbations to  mispredict poses. Furthermore, we demonstrate how these models can be fooled into predicting specific incorrect target poses, indicating their vulnerabilities.
Particularly, pose estimation models are notably vulnerable to targeted attack on the yaw. However, it is relatively harder to fool the model into predicting that the vehicle is moving backwards.
We also note that attacking the translation affects the rotation predictions and vice versa. However, adversarial perturbations that attack pose differ from those that attack depth, even though both models are trained together in an unsupervised manner.
Therefore, adversarial perturbations that fool the pose estimation models do not fool the depth estimation models, and vice versa.  
Finally,
we show that adversarial images generated using one pose model can be used to also attack other pose models with different architectures and training losses. 
Therefore, our work sheds light on the issues concerning reliable deployment of pose estimation models. 
Consequently, future works should focus on defense mechanisms against such attacks, especially for safety-critical applications.

\addtolength{\textheight}{-12cm}   


\bibliographystyle{IEEEtran}
\bibliography{IEEEabrv,ref}

\begin{thebibliography}{10}
\providecommand{\url}[1]{#1}
\csname url@rmstyle\endcsname
\providecommand{\newblock}{\relax}
\providecommand{\bibinfo}[2]{#2}
\providecommand\BIBentrySTDinterwordspacing{\spaceskip=0pt\relax}
\providecommand\BIBentryALTinterwordstretchfactor{4}
\providecommand\BIBentryALTinterwordspacing{\spaceskip=\fontdimen2\font plus
\BIBentryALTinterwordstretchfactor\fontdimen3\font minus
  \fontdimen4\font\relax}
\providecommand\BIBforeignlanguage[2]{{%
\expandafter\ifx\csname l@#1\endcsname\relax
\typeout{** WARNING: IEEEtran.bst: No hyphenation pattern has been}%
\typeout{** loaded for the language `#1'. Using the pattern for}%
\typeout{** the default language instead.}%
\else
\language=\csname l@#1\endcsname
\fi
#2}}

\bibitem{packnet}
V.~Guizilini, \emph{et~al.}, ``3d packing for self-supervised monocular depth
  estimation,'' in \emph{CVPR}, 2020.

\bibitem{bian2021unsupervised}
J.~W. Bian, \emph{et~al.}, ``Unsupervised scale-consistent depth learning from
  video,'' \emph{ICCV}, 2021.

\bibitem{varma2022transformers}
A.~Varma., H.~Chawla., B.~Zonooz., and E.~Arani., ``Transformers in
  self-supervised monocular depth estimation with unknown camera intrinsics,''
  in \emph{VISAPP}, 2022.

\bibitem{pillai2017towards}
S.~Pillai and J.~J. Leonard, ``Towards visual ego-motion learning in robots,''
  in \emph{IROS}, 2017.

\bibitem{huang2021self}
B.~Huang, \emph{et~al.}, ``Self-supervised generative adversarial network for
  depth estimation in laparoscopic images,'' in \emph{MICCAI}, 2021.

\bibitem{kurakin2017adversarial}
A.~Kurakin, I.~Goodfellow, and S.~Bengio, ``Adversarial machine learning at
  scale,'' in \emph{ICLR}, 2017.

\bibitem{yuan2019adversarial}
X.~Yuan, P.~He, Q.~Zhu, and X.~Li, ``Adversarial examples: Attacks and defenses
  for deep learning,'' \emph{IEEE Transactions on Neural Networks and Learning
  Systems}, 2019.

\bibitem{hendrik2017universal}
J.~Hendrik~Metzen, M.~Chaithanya~Kumar, T.~Brox, and V.~Fischer, ``Universal
  adversarial perturbations against semantic image segmentation,'' in
  \emph{ICCV}, 2017.

\bibitem{arnab2018robustness}
A.~Arnab, O.~Miksik, and P.~H. Torr, ``On the robustness of semantic
  segmentation models to adversarial attacks,'' in \emph{CVPR}, 2018.

\bibitem{qin2019adversarial}
C.~Qin, \emph{et~al.}, ``Adversarial robustness through local linearization,''
  \emph{NeurIPS}, 2019.

\bibitem{tramer2019adversarial}
F.~Tramer and D.~Boneh, ``Adversarial training and robustness for multiple
  perturbations,'' \emph{NeurIPS}, 2019.

\bibitem{zhang2020adversarial}
Z.~Zhang, \emph{et~al.}, ``Adversarial attacks on monocular depth estimation,''
  \emph{arXiv preprint arXiv:2003.10315}, 2020.

\bibitem{wong2020targeted}
A.~Wong, S.~Cicek, and S.~Soatto, ``Targeted adversarial perturbations for
  monocular depth prediction,'' in \emph{NeurIPS}, 2020.

\bibitem{li2019pose}
Y.~Li, Y.~Ushiku, and T.~Harada, ``Pose graph optimization for unsupervised
  monocular visual odometry,'' in \emph{ICRA}, 2019.

\bibitem{DBLP:conf/iclr/MadryMSTV18}
A.~Madry, \emph{et~al.}, ``Towards deep learning models resistant to
  adversarial attacks,'' in \emph{ICLR}, 2018.

\bibitem{DBLP:journals/corr/SzegedyZSBEGF13}
C.~Szegedy, \emph{et~al.}, ``Intriguing properties of neural networks,'' in
  \emph{ICLR}, 2014.

\bibitem{DBLP:journals/corr/GoodfellowSS14}
I.~J. Goodfellow, J.~Shlens, and C.~Szegedy, ``Explaining and harnessing
  adversarial examples,'' in \emph{ICLR}, 2015.

\bibitem{xie2017adversarial}
C.~Xie, \emph{et~al.}, ``Adversarial examples for semantic segmentation and
  object detection,'' in \emph{ICCV}, 2017.

\bibitem{zhou2017unsupervised}
T.~Zhou, M.~Brown, N.~Snavely, and D.~G. Lowe, ``Unsupervised learning of depth
  and ego-motion from video,'' in \emph{CVPR}, 2017.

\bibitem{Geiger2012CVPR}
A.~Geiger, P.~Lenz, and R.~Urtasun, ``Are we ready for autonomous driving? the
  kitti vision benchmark suite,'' in \emph{CVPR}, 2012.

\bibitem{he2016deep}
K.~He, X.~Zhang, S.~Ren, and J.~Sun, ``Deep residual learning for image
  recognition,'' in \emph{CVPR}, 2016.

\bibitem{sturm2012evaluating}
J.~Sturm, W.~Burgard, and D.~Cremers, ``Evaluating egomotion and
  structure-from-motion approaches using the tum rgb-d benchmark,'' in
  \emph{Workshop on Color-Depth Camera Fusion in Robotics at IROS}, 2012.

\bibitem{chawla2021multimodal}
H.~Chawla, A.~Varma, E.~Arani, and B.~Zonooz, ``Multimodal scale consistency
  and awareness for monocular self-supervised depth estimation,'' in
  \emph{ICRA}, 2021.

\end{thebibliography}
\end{document}